\newif\ifarXiv         
\newif\ifjournal        

 \journalfalse \arXivtrue  

\ifjournal 
\input{macro_journal}
\fi

\ifarXiv     
\documentclass[11pt]{article}  
\usepackage{amssymb}    
\usepackage{graphicx}          
\usepackage{amsmath}         
\usepackage{caption}
\usepackage{subcaption}
\usepackage{dsfont}
\usepackage{listings}
\usepackage{titlepic}
\usepackage{xcolor}
\usepackage{algorithm,algorithmicx}
\usepackage{todonotes}
\usepackage{array}
\usepackage{booktabs}
\usepackage{bm}
\usepackage{fullpage}

\numberwithin{equation}{section}
\allowdisplaybreaks[3]

\usepackage{hyperref}
\hypersetup{
    colorlinks=true,
    linkcolor=blue,
    filecolor=magenta,      
    urlcolor=cyan,
    citecolor = blue,
}
\usepackage{ dsfont }
\usepackage{listings}
\usepackage{bbm}
\usepackage{color}
\usepackage{ dsfont }
\usepackage{enumerate}
\usepackage{todonotes}

\usepackage{booktabs}  

\def\by{{y}}

\def\R{\mathbb{R}}

\newcommand{\Ebracket}[1]{\mathbb{E}\left[{#1}\right]}
\def\E{\mathbb{E}}
\newcommand{\rbracket}[1]{\left(#1\right)}

\newcommand{\norm}[1]{\left\|#1\right\|}

\newcommand{\innerp}[1]{\langle{#1}\rangle}

\newcommand{\argmin}[1]{\underset{#1}{\operatorname{arg}\operatorname{min}}\;}

\def\calE{\mathcal{E}}
\def\calN{\mathcal{N}}

\def\Gbar{ {\overline{G}} }

\def\LGbar{ \mathcal{L}_{\Gbar}}  

\def\calS{\mathcal{S}}
\def\calT{\mathcal{T}}

\def\spaceY{\mathbb{Y}}
\def\spaceYmu{{L^2_\mu(\mathcal{T})}}

\newcommand{\beqa}{\begin{eqnarray}}
\newcommand{\eeqa}{\end{eqnarray}}
\newcommand{\beqas}{\begin{eqnarray*}}
\newcommand{\eeqas}{\end{eqnarray*}}
\newcommand{\beq}{\begin{equation}}
\newcommand{\eeq}{\end{equation}}
\newcommand{\beqs}{\begin{equation*}}
\newcommand{\eeqs}{\end{equation*}}

\newtheorem{theorem}{Theorem}

\newtheorem{assumption}[theorem]{Assumption}

\newtheorem{definition}[theorem]{Definition}

\newtheorem{lemma}[theorem]{Lemma}
\newtheorem{proposition}[theorem]{Proposition}
\newtheorem{remark}[theorem]{Remark}

\newenvironment{proof}[1][Proof]{\noindent\textbf{#1.} }{\ \rule{0.5em}{0.5em}}

\numberwithin{equation}{section}
\numberwithin{theorem}{section}

\usepackage{xcolor}
\definecolor{darkmagenta}{rgb}{0.55, 0.0, 0.55}

\title{Small noise analysis for Tikhonov and RKHS regularizations}
\bigskip
\author{ Quanjun Lang \thanks{Department of Mathematics, Duke University,  Durham, NC,  27707, USA
(\texttt{\textcolor{black}{quanjun.lang@duke.edu}}) }
  \and Fei Lu \thanks{Department of Mathematics, Johns Hopkins University, Baltimore, MD 21218, USA
(\texttt{\textcolor{black}{feilu@math.jhu.edu}}) }
  }
  
\fi

\let\OLDthebibliography\thebibliography
\renewcommand\thebibliography[1]{
  \OLDthebibliography{#1}
  \setlength{\parskip}{0pt}
  \setlength{\itemsep}{3pt}
}

\usepackage{tabularx}
\usepackage{enumitem}
\usepackage{cases}

\begin{document}

\maketitle

\begin{abstract}
 Regularization plays a pivotal role in ill-posed machine learning and inverse problems. However, the fundamental comparative analysis of various regularization norms remains open. We establish a small noise analysis framework to assess the effects of norms in Tikhonov and RKHS regularizations, in the context of ill-posed linear inverse problems with Gaussian noise. This framework studies the convergence rates of regularized estimators in the small noise limit and reveals the potential instability of the conventional $L^2$-regularizer. We solve such instability by proposing an innovative class of adaptive fractional Sobolev space regularizers, which covers the $L^2$ Tikhonov and RKHS regularizations by adjusting the fractional smoothness parameter. A surprising insight is that over-smoothing via these fractional Sobolev spaces consistently yields optimal convergence rates, but the optimal hyper-parameter may decay too fast to be stably selected in practice. 
\end{abstract}

  \ifjournal  
\begin{keywords}
  adaptive RKHS, regularization, small noise, ill-posed inverse problem \end{keywords}
\fi

\ifarXiv
\textbf{Keywords}: ill-posed inverse problem, adaptive RKHS, Tikhonov regularization, small noise  \\
\fi 


 
\section{Introduction}
Consider the ill-posed inverse problem of minimizing a quadratic loss function
\begin{equation}
	\min_{\phi\in L^2_\rho} \mathcal{E}(\phi):=
	\frac{1}{n}\sum_{k=1}^n\norm{L_k\phi-\by_k}^2_{\spaceY}  
\end{equation}
where $L_k:L^2_\rho\to \spaceY$ are bounded linear operators between the Hilbert spaces $L^2_\rho$ and $\spaceY$. This problem is ill-posed in the sense that the mini-norm least squares solution is sensitive to noise in data or approximation error in computation. Such ill-posedness is common in inverse problems such as the Fredholm integral equations and nonparametric regression in statistical and machine learning.  

Regularization is crucial in solving the above ill-posed inverse problems. Numerous regularization methods address this long-standing challenge. Roughly speaking, these methods restrict the function space of optimization search in two approaches: (i) setting bounds on the function or the loss functional (e.g., minimizing $\calE(\phi)$ subject to $\|\phi\|_*<\delta$ or minimizing $\|\phi\|_*$ subject to $\calE(\phi)<\alpha$), and (ii) penalizing the loss function by a square norm as in the well-known Tikhonov regulation (see e.g., \cite{tihonov1963solution,tikhonov1995numerical,hansen1998rank}): 
\begin{equation*}
\widehat \phi_\lambda = \argmin{\phi} \calE_\lambda(\phi) : =  \mathcal{E}(\phi)+\lambda \norm{\phi}_{*}^2.   
\end{equation*}
Here, $\|\phi\|_*$ is a regularization norm, and $\delta,\alpha, \lambda$ are the hyper-parameters controlling the strength of regularization. Given a norm, the minimization is often solved by either direct matrix decomposition-based methods to select an optimal hyper-parameter as in  \cite{wahba1977practical,hansen1998rank,belge2002efficient} or iterative methods as in  \cite{gazzola2019ir,chen2022stochastic,zhang2022stochastic}. Meanwhile, various norms have been proposed, including the commonly-used $L^2$ norm, the Sobolev norms in  \cite{rudin1992nonlinear}, and the RKHS norms (see e.g., \cite{smale2007learning,bauer2007regularization,CZ07book,LLA22}). 

However, the fundamental comparative analysis of the regularization norms remains open. Such a comparison is challenging as it requires extracting essentials from multiple factors, including the random noise, the spectrum of the normal operator, the smoothness of the solution, the model error, the selection of the hyper-parameter, and the method of optimization.  

\begin{figure}\ifjournal \vspace{-3mm} \fi
    \centering
    \includegraphics[width=0.9\textwidth]{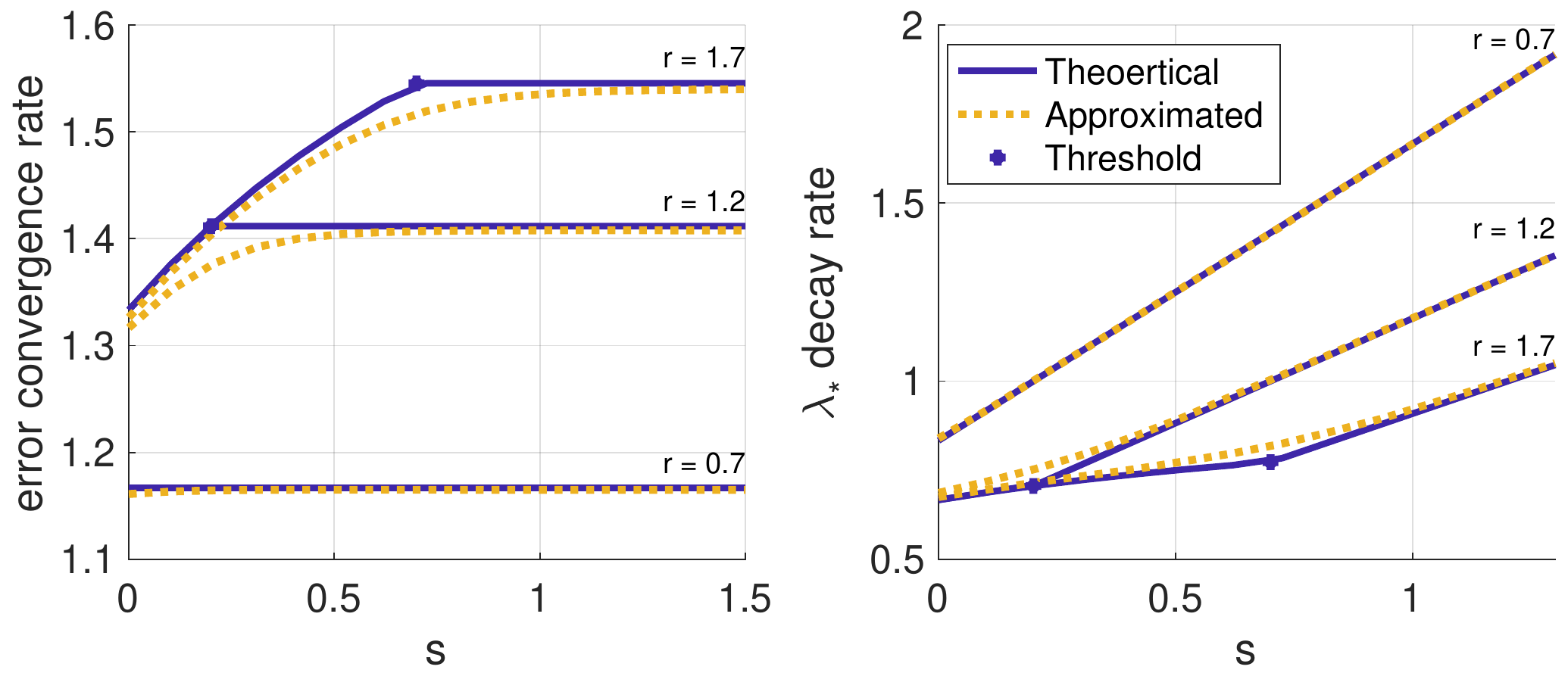}
    \caption{over-smoothing ($s>r - \frac{\beta+1}{2}$) yields the optimal convergence rate (left), at the price of fast decaying optimal hyper-parameter $\lambda_*$ (right). Here $s$ is the smoothness parameter of the fractional Sobolev space for regularization, $r$ is the regularity of the true function, and $\beta$ represents the spectral decay of the normal operator. See Sect.\ref{sec:oracle-rate} and Theorem \ref{thm:snl_reguEst} for details. 
    } 
     \label{fig:thm_apx} \ifjournal \vspace{-3mm} \fi
\end{figure}

We introduce a small noise analysis framework making this comparison possible for Tikhonov regularization. This framework compares the convergence rates of the regularized estimators with the oracle optimal hyper-parameters in the small noise limit. It generalizes the classical idea of the bias-variance trade-off in learning theory (see, e.g., \cite{CuckerSmale02,CZ07book}) to inverse problems. Our main contributions are threefold.  
\begin{itemize}[leftmargin=6mm]  
	\item We introduce a new \emph{small noise analysis} framework to study the effects of norms in Tikhonov and RKHS regularizations. This framework extends the classical idea of bias-variance trade-off in learning theory to inverse problems in the small noise limits. 
	As far as we know, this is the first study enabling a theoretical comparison between regularization norms through convergence rates in small noise.     
	\item We introduce a new class of \emph{adaptive fractional Sobolev spaces} $H_G^s$ with a smoothness parameter $s\geq 0$ for regularization; see Definition \ref{def:H-s}. It recovers the $L^2$-norm when $s=0$ and an RKHS norm in DARTR in \cite{LLA22} when $s=1$.  This regularization restricts the minimization to search within the function space of identifiability, leading to stable regularized estimators in the small noise limit.      
	\item We show that the \emph{optimal convergence rates} of $H_G^s$-regularized estimators depend on the smoothness parameter $s$, the regularity of the true function, and the spectrum decay of the normal operator. Our analysis shows a surprising insight that over-smoothing retains the optimal rate of convergence, but the optimal hyper-parameter may decay too fast for computational practice; see Fig.~\ref{fig:thm_apx} for decay rates of optimal parameters and see Fig.~\ref{fig:FIE} for instability of hyper-parameters selected in practice. Thus, we recommend replacing the conventional $L^2$-regularizer with a fractional Sobolev space regularizer with a medium level of smoothing.  
\end{itemize}

\paragraph{Notations.} The notation $X\sim \calN(a,\Sigma)$ means that the random variable $X$ has a Gaussian distribution with mean $a$ and covariance $\Sigma$. $L^2_\rho$ denotes the function space of square-integrable functions defined on the support of the measure $\rho$. We denote by $\LGbar^{-1}$ the pseudo-inverse of the operator $\LGbar$. We denote by $\calE$ the loss function,$\lambda$ the hyper-parameter controlling the regularization strength.    

\ifjournal \vspace{-3mm} \fi
\paragraph{Related works.}
A large literature has studied \emph{convergence of regularized estimators} in inverse problems and in statistical learning, and we mention only those close to our setting, particularly those using explicitly the error norm to select the optimal hyper-parameter as in \cite{hansen1998rank}. In the case of vanishing measurement error, 
convergence has been proven in \cite[Theorem 1.1, page 8]{tikhonov1995numerical} without a rate;
when the measurement error decays uniformly, the convergence rate of the $L^2$-regularized estimator has been studied in \cite[Theorem 5.8, page 131]{engl1996regularization}. 
Also, convergence rates with refining mesh have been studied in \cite{wahba1973convergence,wahba1977practical,wen2011regularized,nashed1974generalized,Li2005modified}. 

On the other hand, \emph{RKHS regularization} has been widely studied in inverse problems  \cite{wahba1977practical,nashed1974generalized,Li2005modified}, functional linear regression (see e.g.,\cite{hall2007methodology,yuan_cai2010,balasubramanian2022unified}), learning theory (see e.g.,\cite{CuckerSmale02,smale2007learning,CZ07book,bauer2007regularization,chada2022data,scampicchio2023error}) using pre-specified reproducing kernels, where the inverse problem in the large sample limit is a well-posed. Additionally, small noise analysis has been used to study maximum a posterior of Bayesian inverse problems in \cite{dashti2013map,chada2022data}, samplers in  \cite{goodman2016small}, and numerical solutions to stochastic differential equations in \cite{buckwar2010stochastic}.  

However, none of these studies considers the adaptive fractional Sobolev space regularization introduced in this study. Neither do they discover the delicate dependence of the optimal convergence rate on the spectrum decay and the regularity of the true solution.

\ifjournal \vspace{-3mm} \fi

\section{Tikhonov and RKHS regularization} \label{sec:FSOI} 

\subsection{Ill-posed linear inverse problems}

Let $\rho$ be a probability measure with compact support $\calS$. Consider the inverse problem: 
\begin{equation}\label{eq:ip}
\text{recover } \phi\, \text{ in }\,	y = L\phi + \sigma \dot W\, \text{ from data }  \{(L_k,y_k)\}_{k=1}^n, 
\end{equation}
 where $L_k: L^2_\rho \to \spaceY$ are bounded linear operators between the Hilbert spaces $L^2_\rho$ and $\spaceY$. Here $\dot W$ is a white noise on $\spaceY$, i.e., $\innerp{\dot W, y}_\spaceY\sim \calN(0,\|y\|_\spaceY^2)$ for each $y\in \spaceY$; see \cite{da2014stochastic} for Gaussian measures on infinite-dimensional spaces. When the space $\spaceY$ is finite-dimensional,  Eq.\eqref{eq:ip} can be viewed as a strong form equation in $\spaceY$ since $\dot W$ can be represented as a standard Gaussian random variable, and when $\spaceY$ is infinite-dimensional, Eq.\eqref{eq:ip} is interpreted in its weak form.  Such inverse problems arise in a wide range of applications, including the Fredholm integral equations of the first kind, learning kernels in nonlocal operators, and deconvolution problems; see Section \ref{sec:examples} for a few examples.

The goal of this study is to investigate the convergence rates of regularized estimators when $\sigma\to0$, where the positive constant $\sigma$ represents the strength of the noise. Thus, throughout this study, the sample size $n$ is fixed.

We focus on ill-posed problems in $L^2_\rho$ that the \emph{mini-norm least squares estimator} (LSE) is sensitive to noise in data or approximation error. That is, in the variational formulation of the inverse problem, the minimal-norm minimizer of the loss function
\beqa \label{eq:lossFn}
&&\mathcal{E}(\phi):= \frac{1}{n}\sum_{k=1}^n \norm{L_k\phi-y_k}^2_{\spaceY} = \frac{1}{n}\sum_{k=1}^n\innerp{L_k^*L_k\phi,\phi}_{L^2_\rho}-2\innerp{L_k^*y_k,\phi}_{L^2_\rho}+\norm{y_k}^2_{\spaceY} 
\label{least_sq_org}
\eeqa
is sensitive to the noise in the data $\{y_k\}$. Here $L^*_k$ denotes the adjoint operator of $L_k$. Such ill-posedness roots in the unboundedness of the (pseudo-)inverse of the data-dependent normal operator 
\begin{equation}\label{eq:LGbar}
\LGbar: = \frac{1}{n}\sum_{k=1}^n L_k^*L_k: L^2_\rho\to L^2_\rho, 
\end{equation}
e.g., when $\LGbar$ is compact with eigenvalues converging to zero (see examples in Sect.~\ref{sec:examples}). 

To be precise, we introduce the following assumptions and definitions. 
Note first that by definition, $\LGbar$ is a semi-positive self-adjoint operator, i.e., $\innerp{\LGbar\phi,\phi} = \innerp{\phi,\LGbar\phi}\geq 0$ for any $\phi\in L^2_\rho$.   
 \begin{assumption}\label{assumption:LG}
 	The operator $\LGbar$ is a trace-class operator, i.e., $\sum_{i}\lambda_i<\infty$, where $\{\lambda_i\}_{i\geq 1}$ denote the positive eigenvalues of $\LGbar$ arranged in a descending order. 
 \end{assumption} 
  Denote $\{\psi_i,\,\psi_j^0\}_{i,j}$ be the orthonormal eigenfunctions of $\LGbar$ with $\psi_i$ and $\psi_j^0$ corresponding to eigenvalues $\lambda_i$ and zero (if any). Then, $\{\lambda_i\}_{i\geq 1}$ is either finite or $\lambda_i\to 0$ as $i\to \infty$, and these eigenfunctions form a complete orthonormal basis of $L^2_\rho$. 
      
We define the function space of identifiability as the subspace of $L^2_\rho$ in which the inverse of $\LGbar$ is well-defined, i.e., the loss function has a unique minimizer. 
\begin{definition}[Function space of identifiability (FSOI)]\label{def:fsoi}
 The FSOI of the inverse problem in \eqref{eq:ip} is $H=\mathrm{Null}(\LGbar)^\perp$, the complement of the null space of the operator $\LGbar$. 
  \end{definition}

\begin{definition}[Ill-posed inverse problem] The inverse problem in \eqref{eq:ip} is ill-posed in $L^2_\rho$ if the operator $\LGbar^{-1}: H\to L^2_\rho$ is unbounded, where $H=\mathrm{Null}(\LGbar)^\perp$ is the FSOI.      \end{definition}

Note that ill-posedness differs from ill-conditionedness. An ill-conditioned problem has a large condition number $\lambda_{max}/\lambda_{min}$, where $\lambda_{max}$ and $\lambda_{min}$ are maximal and minimal positive eigenvalues of $\LGbar$. Thus, an ill-posed problem is always ill-conditioned, but a well-posed inverse problem can be ill-conditioned. Also, when $\LGbar$ is finite-rank (i.e., it has only finitely many nonzero eigenvalues), the inverse problem is well-posed.

The next lemma characterizes the mini-norm LSE; see Appendix \ref{sec:append} for its proof.  
\begin{lemma}[The mini-norm LSE]\label{lemma:LSE}
Under Assumption {\rm\ref{assumption:LG}}, the mini-norm LSE is the unique minimizer of the loss function in \eqref{eq:lossFn} in the space $H$, and it can be written as 
\beq\label{eq:LSE}
 \widehat{\phi} = \LGbar^{-1}\phi^\by, \quad
\phi^\by := \frac{1}{n}\sum_{k = 1}^n L_k^* y_k.  
\eeq
Let $\phi_*$ be the true solution, i.e., $y_k = L_k \phi_*+\sigma \dot W_k$. The following statements hold true. 
\begin{itemize}[leftmargin=6mm]  
\item[{\rm(a)}]  The function $\phi^\by$ has a decomposition 
\begin{equation}\label{eq:noise_dec}
\phi^\by = \LGbar \phi_*+ \phi^\sigma, \, \text{ with } \phi^\sigma= \sum_i \sigma \xi_i \lambda_i^{1/2}\psi_i, 
\end{equation} 
where $\{\xi_i\}$ are independent and identically distributed Gaussian random variables. That is, $\phi^\sigma \sim \mathcal{N}(0, \sigma^2\LGbar)$ with $\Ebracket{\|\phi^\sigma\|_{L^2_\rho}^2}= \sigma^2\sum_i\lambda_i$. 
\item[{\rm(b)}]  When $\sum_{i} \lambda_i^{-1}<\infty$, the estimator $\widehat{\phi} = \LGbar^{-1}\phi^\by 
\sim \calN(\phi_*,\sigma^2\LGbar^{-1})$ is well-defined; but when $\sum_{i} \lambda_i^{-1}=\infty$, the pseudo-inverse $\LGbar^{-1}\phi^\by$ is ill-defined in $L^2_\rho$ since $\Ebracket{\|\LGbar^{-1}\phi^\by\|_{L^2_\rho}^2} = \infty$. 
\item[{\rm(c)}]  When the data is noiseless, we have $\widehat{\phi} = \LGbar^{-1}\phi^\by= P_H\phi_*$, where $P_H:L^2_\rho\to H$ is the projection operator. 

\end{itemize}
\end{lemma}

The above lemma reveals the nature of the ill-posedness and provides insights into regularization. The inverse problem is ill-defined beyond the FSOI $H$, and it is ill-posedness in $H$ when the positive eigenvalues of $\LGbar$ converge to zero. 
As a result, when regularizing the ill-posed problem, it is important to ensure that the solution lies in the FSOI. 

Before introducing such regularization strategies in Sect.~\ref{sec:RKHS}, we first introduce a few examples of the operator $\LGbar$.        

\subsection{Examples}\label{sec:examples}

As illustrations, we consider integral operators $L_k: L^2_\nu(\calS)\to L^2_\mu(\calT)$ with kernels $K_k$:
\begin{align}\label{eq:L-opterator}
L_k\phi(t) = \int_\calS K_k(t,s)\phi(s) \nu(ds), \, t\in \calT. 
\end{align}
Here $\calS\subset \R$ and $\calT\subset \R$ are two compact sets, which can be either intervals or sets with finitely many elements. The measure $\nu$ is a finite measure with support $\calS$: it is the Lebesgue measure when $\calS$ is an interval, and it is an atom measure when $\calS$ has finitely many elements. 
Similarly, we let $\mu$ be a finite measure on $\calT$. Three examples are as follows: 
\begin{itemize}[leftmargin=6mm]  
	\item \emph{Fredholm integral equations} of the first kind, in which one solves $\phi$ from $(K_1,y_1)$:  
	\begin{align}\label{eq:FIE}
y_1(t) = L_1\phi(t) +\sigma \dot W(t) = \int_\calS K_1(t,s)\phi(s) ds +  \sigma \dot W(t), \, t\in \calT, 
\end{align}
where $\calS$ and $\calT$ are closed intervals (for example, $\calS=[0,1]$ and $\calT=[1,2]$) with $\nu$ and $\mu$ being the Lebesgue measure.
It is a prototype of ill-posed inverse problems dating backing from \cite{hadamard1923lectures}, and it is a testbed for regularization methods, see e.g., \cite{wahba1973convergence,hansen1998rank,Li2005modified,LuOu23}. 
\item \emph{Learning kernels in nonlocal operators}, in which one aims to recover the kernel $\phi$ in the nonlocal operator $R_\phi:H^1_{0,\pi}\to \spaceY$ with $H^1_{0,\pi}=\{u:[0,2\pi]\to \R: u, u'\in ^2([0,2\pi]), 2\pi\text{-periodic}\}$:   
	\begin{align*}
y_k(t) = R_\phi[u_k] +\sigma \dot W_k(t), L_k\phi(t)  =R_\phi[u_k]= \int_\calS K_k(t,s)\phi(s) ds, \, t\in \calT 
\end{align*}
from data pairs $\{(u_k,y_k)\}$. Here $\calS=[0,2\pi]=\calT$, and $K_k$ can be either $K_k(t,s) = \partial_t [u_k(t-s)u_k(t)- u_k(t+s)u_k(t)]$ in \cite{LangLu22,LAY22} or $K_k(t,s) =  u_k(t+s)+u_k(t-s)-2u_k(t)$ in \cite{youYu2022_Datadriven}. 
\item \emph{Recovering kernels in Toeplitz matrix or Hankel matrix} from data (see, e.g.,\cite{Gray06}).  
\end{itemize}

In these examples, the measure $\rho$ can be either the Lebesgue measure on its support or the data-dependent measure 
$\frac{d\rho}{d\nu}(s):=\frac{1}{Z} \sum_{k=1}^n\int_\calT | K_i(t,s)| \mu(dt), \, \forall s\in \calS
$, 
where $Z$ is a normalizing constant.  
The operator $\LGbar$ in \eqref{eq:LGbar} is determined by the bilinear form 
$ \innerp{\LGbar\phi,\psi} = \frac{1}{n}\sum_{k=1}^n \innerp{L_k\phi,L_k\psi}_{\spaceYmu}= \int_\calS \int_\calS \phi(s)\psi(s')\Gbar(s,s')\rho(ds) \rho(ds') 
$,
where the bivariate function $\Gbar:\calS\times \calS \to \R$ is defined as
 \beq 
  \Gbar(s,s'):= \frac{1}{\frac{d\rho}{d\nu}(s)\frac{d\rho}{d\nu}(s')}\int_\calT \frac{1}{n}\sum_{k=1}^n  K_k(t,s)K_k(t,s') \mu(dt). \notag \eeq
In fact, $\LGbar$ is an integral operator with $\Gbar$ as its kernel, i.e.
$ \LGbar\phi(r):=\int_a^b \phi(s)\Gbar(r,s)\rho(ds)
$. 
The function $\Gbar$ is a Mercer kernel when the functions $K_k: \calT\times \calS \to \R$ are continuous. Hence, it defines an RKHS, $H_G = \LGbar^{1/2}(L^2_\rho)$. The FSOI is the closure of this RKHS in $L^2_\rho$ (see \cite{LLA22}).

\subsection{Adaptive fractional Sobolev spaces and  regularization}\label{sec:RKHS}
  
We introduce a class of adaptive fractional Sobolev spaces for regularization that can ensure the optimization search takes place inside the data-dependent FSOI. These RKHSs are constructed through the eigen-decomposition of the operator $\LGbar$, having a parameter $s$ controlling the smoothness.  
\begin{definition}[Adaptive fractional Sobolev spaces] \label{def:H-s}
For $s\geq0$, the $s$-fractional Sobolev space associated with the operator $\LGbar$ in $L^2_\rho$ is $H^s_G=\LGbar^{s/2}(L^2_\rho)$, with norm $\|\phi\|_{H^s_G}^2 = \|\LGbar^{-s/2} \phi\|_{L^2_\rho}^2$ for all $\phi = \sum_{i: \lambda_i>0} c_{i} \psi_i$ such that $\sum_{i} \lambda_i^{-s}  c_{i}^2<\infty$, 
  where $\{(\lambda_i,\psi_i)\}$ are the positive eigenvalue and eigenfunction pairs of $\LGbar$. 
  \end{definition}     
The fractional Sobolev space recovers the RKHS $H_G$ with reproducing kernel $\Gbar$ when $s=1$. Also, when $s=0$, the fractional Sobolev space is $H_G^0= H$, the FSOI. In general, the larger $s$ is, the smoother are the functions in $H_G^s$; and for all $s>1$,   
\[
H^s_G \subset H_G^1= H_G \subset H_G^0= H\subset L^2_\rho. 
\]

Now we use the norms of these fractional Sobolev spaces  for regularization:
 \begin{equation}\label{eq:reguHs}
\widehat \phi_\lambda^{s} = \argmin{\phi \in H_G^s} \calE_\lambda(\phi) : =  \mathcal{E}(\phi)+\lambda \norm{\phi}_{H_G^s}^2.  
\end{equation}
The next proposition presents the $H_G^s$-regularized estimator and its $L^2_\rho$ error. 
\begin{proposition}[Fractional Sobolev space regularized estimator] \label{prop_error}
The minimizer in \eqref{eq:reguHs} is 
\beqa
\widehat{\phi}_\lambda^{s} 
= (\LGbar+\lambda \LGbar^{-s})^{-1}\phi^\by = (\LGbar^{1+s}+\lambda I_H)^{-1}\LGbar^{s} \phi^\by, \label{eq:est-reguHs}
\eeqa
where $I_H$ is the identity operator on $H$. 
Let the true function be $\phi_*= \sum_i c_i\psi_i + \sum_j d_j\psi_j^0$, where $\{\psi_i,\,\psi_j^0\}_{i,j}$ denote the orthonormal eigenfunctions of $\LGbar$ with $\psi_i$ and $\psi_j^0$ corresponding to eigenvalues $\lambda_i$ and zero (if any). Then, for any $\lambda>0$, the error of this regularized estimator is  
 \beq \label{eq:err-Hs}
   \|\widehat{\phi}_\lambda^{s}-\phi_*\|^2_{L^2_\rho} =   \sum_i (\lambda_i +\lambda \lambda_i^{-s})^{-2}(\sigma \lambda_i^{1/2}\xi_i - \lambda c_i)^2+ \sum_jd_j^2,  
  \eeq
  where $\xi_i$ are iid standard Gaussian. In particular, when $\sum_j d_j^2=0$, the expectation of the error is 
\begin{equation}\label{eq:mse_s}
  \begin{aligned}  
  e(\lambda;s) = &\E \| \widehat{\phi}_\lambda-\phi_*\|^2_{L^2_\rho} = \sum_i (\lambda_i^{s + 1} +\lambda)^{-2}(\sigma^2 \lambda_i^{2s+1} + \lambda^2 c_i^2). 
\end{aligned}
\end{equation}  
\end{proposition}
\begin{proof}
The uniqueness of the minimizers and their explicit form follow from the Fr\'echet derivatives of the regularized loss function. In fact, note that $ \|
\phi\|_{H_G^s}^2 = \innerp{\LGbar^{-s}\phi,\phi}_{L^2_\rho}$ since the operator $\LGbar^{s/2}$ is self-adjoint. Then, similar to \eqref{eq:Frechet},  the Fr\'echet derivative of $\calE_\lambda$ at $\phi\in H_G^s$ under the $L^2_\rho$-topology is  
\[
\nabla \calE_\lambda(\phi) = 2[ (\LGbar + \lambda \LGbar^{-s}) \phi -\phi^\by ]. 
\]
Then, its unique zero in $H_G^s$ is $\widehat{\phi}_\lambda^{s} = (\LGbar + \lambda \LGbar^{-s})^{-1}\phi^\by$. 

To obtain the eigenvalue characterizations of the error, recall the decomposition $\phi^\by = \LGbar \phi_*+\phi^\sigma$ in \eqref{eq:noise_dec}. Then, we obtain \eqref{eq:err-Hs}
 by noticing that 
\begin{align}
 \widehat{\phi}_\lambda^{s}&=(\LGbar+\lambda \LGbar^{-s})^{-1}(\LGbar{\phi_*}+\lambda \LGbar^{-s} \phi_*-\lambda \LGbar^{-s} \phi_*+\phi^\sigma) \notag \\
&=\phi_*+(\LGbar+\lambda \LGbar^{-s})^{-1}(-\lambda\LGbar^{-s} \phi_*+\phi^\sigma).  \label{eq:error-decomp}
\end{align}
Additionally, using the fact that $\phi^\sigma= \sum_i \sigma \lambda_i^{1/2}\xi_i\psi_i$ with $\xi_i$ being iid standard Gaussian, we obtain \eqref{eq:mse_s} by taking expectation.  
\end{proof}
 
Note that when $s=0$, this regularization reduces to the conventional $L^2$-Tikhonov regularization but with the minimization search constrained inside the FSOI. When $s=1$, it recovers the DARTR in \cite{LLA22}.

Proposition \ref{prop_error} demonstrates the complexity of choosing an optimal hyper-parameter $\lambda$. An optimal $\lambda$ minimizing the error in \eqref{eq:err-Hs} must balance the error caused by the noise and the bias caused by the regularization. It depends on the spectrum of the operator, the true solution (which may have non-identifiable components outside the FSOI), and the realization of the noise. In particular, the dependence on the realization of the noise makes the analysis of the regularized estimator intractable. Thus, we consider the expectation of the error in \eqref{eq:mse_s} in our small noise analysis in the next section, and seek an optimal $\lambda$ to achieve a tradeoff between the bias $\E\| (\LGbar+\lambda \LGbar^{-s})^{-1}(-\lambda \phi_*)\|_{L^2_\rho}^2$ caused by regularization and the variance $\E\| (\LGbar+\lambda \LGbar^{-s})^{-1}(\phi^\sigma)\|_{L^2_\rho}^2$ caused by the noise.  
 
 
\section{Small noise analysis for regularized estimators}\label{sec:conv} 
We introduce a small noise analysis framework that studies the convergence of the regularized estimator in the small noise limit. We show first that the fractional Sobolev space regularizer removes the bias outside the FSOI, thus leading to estimators with vanishing errors, whereas the conventional $L^2$ Tikhonov regularizer cannot. Thus, when studying convergence rates, we don't consider the conventional $L^2$-regularizer and only compare the fractional Sobolev space regularizers with different smoothness parameters $s$. We will show that over-smoothing by the fractional Sobolev spaces always leads to optimal convergence rates.    

\subsection{Finite-rank operators: removing bias outside the FSOI}
We show that the conventional $L^2$ Tikhonov regularizer has the drawback of not removing the perturbation outside the FSOI (often caused by numerical error and/or model error), leading to an estimator failing to converge in the small noise limit. In contrast, the fractional Sobolev space regularizer removes the bias outside the FSOI, consistently leading to convergent estimators.  
 \begin{proposition}
 Assume that the operator is finite rank, the true function is inside the FSOI, and a presence of perturbations outside the FSOI as follows: 
 \begin{itemize}[leftmargin=6mm]  
 \item the operator $\LGbar$ has descending eigenvalues satisfying $\lambda_i=0$ for all $i>K$ and $\lambda_K>0$; 
 \item the term $\phi^\by$ in \eqref{eq:LSE} is approximated by $\widetilde\phi^\by$ such that $\widetilde\phi^\by = \phi^\by + \phi^\epsilon$ with $\phi^\epsilon\in \mathrm{null}(\LGbar)$; 
 \item the true function $\phi_*= \sum_{1\leq i\leq K} c_i\psi_i$, where $\{\psi_i\}$ are the eigen-functions of $\{\lambda_i\}$.   	
 \end{itemize}
	Then, when $\sigma\to 0$, the conventional $L^2$-regularized estimator $\widetilde{\phi}_\lambda^{L^2}= (\LGbar+\lambda I)^{-1} \widetilde \phi^\by$ has a non-vanishing bias; in contrast, the fractional Sobolev space regularized estimator $\widetilde{\phi}_\lambda^{s}= (\LGbar+\lambda \LGbar^{-s})^{-1} \widetilde \phi^\by$ has a vanishing bias.      
 \end{proposition}
 \begin{proof}
 	Let $\phi^\epsilon = \sum_{i>K} \epsilon_i\psi_i$ with $0<\|\phi^\epsilon\|_{L^2_\rho}^2= \sum_{i>K}\epsilon_i^2<\infty$. Then, applying the decomposition \eqref{eq:error-decomp}, we have  
 	\begin{align*}
 	\E[\|\widetilde{\phi}_\lambda^{L^2}-\phi_*\|_{L^2_\rho}^2] &= \sum_{1\leq i\leq K} (\lambda_i +\lambda)^{-2}(\sigma^2 \lambda_i + \lambda^2 c_i^2) + \lambda^{-2}\sum_{i> K} \epsilon_i^2. 
 	\end{align*}
 	Using the fact that $(\lambda_i +\lambda)^{-2}(\sigma^2 \lambda_i + \lambda^2 c_i^2) \geq (\lambda_i + \lambda)^{-2}\lambda^2  c_i^2 \geq \lambda_K^{-2}\lambda^2  c_i^2$, we obtain
 	\[
 	\E[\|\widetilde{\phi}_\lambda^{L^2}-\phi_*\|_{L^2_\rho}^2] \geq \lambda^2 K\lambda_K^{-2}  \sum_{1\leq i\leq K}  c_i^2 + \lambda^{-2}\|\phi^\epsilon\|_{L^2_\rho}^2 \geq 2\sqrt{K}\lambda_K^{-1} \|\phi_*\|_{L^2_\rho}\|\phi^\epsilon\|_{L^2_\rho}, 
 	\]
 	where the last inequality follows from that fact that $\min_{x>0} ax+x^{-1}b \geq 2\sqrt{ab}$. Hence, the conventional $L^2$-regularizer's estimator has a non-vanishing bias. 
 	
 	Similarly, the $H_G^s$-regularizer's estimator has an expected bias: 
 	 	\[
	\E[\|\widetilde{\phi}_\lambda^{s}-\phi_*\|_{L^2_\rho}^2]=\sum_{1\leq i\leq K} (\lambda_i^{s + 1} +\lambda)^{-2}(\sigma^2 \lambda_i^{2s+1} + \lambda^2 c_i^2). 
 	\] 
 	Hence, the minimal bias is less than its value with $\lambda= \sigma$:
 \begin{align*}
 & \min_{\lambda>0}\E[\|\widetilde{\phi}_\lambda^{L^2}-\phi_*\|_{L^2_\rho}^2] \leq  \E[\|\widetilde{\phi}_{\sigma}^{s}-\phi_*\|_{L^2_\rho}^2] \\
  =& \sum_{1\leq i\leq K} (\lambda_i^{s + 1} +\sigma)^{-2}\sigma^2 ( \lambda_i^{2s+1} +  c_i^2)
  \leq \sigma^2 \lambda_K^{-2s-2} (K  \lambda_1^{2s+1} + \|\phi_*\|_{L^2_\rho}^2). 
 \end{align*}
Consequently, this bias vanishes as $\sigma\to 0$. 	
 \end{proof}

 
\subsection{Infinite-rank operator: convergence rate}
  
We analyze the convergence rate of the regularized estimators as $\sigma\downarrow 0$. The focus is on the dependence of the rate on the smoothness parameter $s$ of $H_G^s$, the regularity of the true function, and the spectrum decay of the normal operator. 

\begin{assumption}[Spectrum decay and smoothness of the true function]
\label{assump:spectral_decay}
The normal operator $\LGbar$ has exponential or polynomial spectrum decay, and the true function is $r$-smooth in the following sense. 
\begin{itemize}[leftmargin=+6mm]  
    \item \emph{(Spectrum decay)}. The eigenvalues of $\LGbar$ decay either exponentially or polynomially: 
    \begin{equation}\label{eq:decay_f}
    \lambda_i = p_i^{-1}f(i) , 
\end{equation}
where $f$ is either exponential, $f(x) = e^{-\theta (x - 1)}$ with $\theta > 0$, or polynomial, $f(x) = x^{-\theta}$ with $\theta > 1$. Here $\{p_i\}$ are perturbations satisfying $0 < a \leq p_i^{-1} \leq b < \infty$ for each $i$. 
\item \emph{(r-smoothness of $\phi_*$)}. Let $\phi_*= \sum_i c_i\psi_i \in  H_G^{r}\subset  L^2_\rho$ be $r$-smooth with respect to the spectrum in the sense that $$|c_i| = \widetilde{p}_i\lambda_i^r,$$ where $r>0$ when $f$ is exponential and $r > \frac{1}{2\theta}$ when $f$ is polynomial. Here $\{\widetilde{p}_i\}$ are perturbations satisfying $0 < a \leq \widetilde{p}_i^{-1} \leq b < \infty$ for each $i$. 
\end{itemize}
\end{assumption}

We define a constant $\beta$ to unify notation for both exponential and polynomial spectrum decay: 
\beq \label{eq:beta}
\beta = 
\begin{cases}
1, & \text{ if } f(x)= e^{-\theta(x-1)}; \\
\theta^{-1} + 1,  &\text{ if } f(x)= x^{-\theta}.
\end{cases}
\eeq
In the computation of the technical Lemma \ref{lem:major_series}, we will compute $\rbracket{f^{-1}}'$. If $f(x) = e^{-\theta(x-1)}$, then $f^{-1}(y) = -\frac{1}{\theta}\ln(y)+1$, so that $\rbracket{f^{-1}}'(y) = -\frac{1}{\theta}y^{-1}$. If $f(x) = x^{-\theta}$, then $f^{-1}(y) = y^{-1/\theta}$, so that $\rbracket{f^{-1}}'(y) = -\frac{1}{\theta}y^{-\frac{1}{\theta}-1}$. Overall, we could unify the two cases by writting $\rbracket{f^{-1}}'(y) = -\frac{1}{\theta} y^{-\beta}$, where $\beta = 1$ for exponential decaying spectrum, or $\beta = \frac{1}{\theta} + 1$ for the power decaying one. In both cases, we have $\beta \geq 1$. The range for $r$ is required so that $\phi_*$ is in $L^2_\rho$, and it can be unified by requiring $r > \frac{\beta - 1}{2}.$

Our main results are the optimal convergence rates of $H_G^s$- regularized estimators in the small noise limit. 
\begin{theorem}[Convergence rates of regularized estimators]\label{thm:snl_reguEst}
Suppose Assumption {\rm\ref{assump:spectral_decay}} holds and recall the $H_G^s$-regularizer's error $e(\lambda,s)$ in \eqref{eq:mse_s}. Then, as $\sigma \to 0$, the optimal parameter $\lambda_*= \argmin{\lambda>0} e(\lambda;s) $ and the estimator's error converge at the rates: 
\begin{equation}\label{eq:opt_lambda}
\lambda_{*}\simeq \begin{cases}
\sigma^{\frac{2s+2}{2r+1}}, & s > r - \frac{\beta + 1}{2},\\
\sigma^{\frac{2s+2}{2s+2+\beta}}, &s < r - \frac{\beta + 1}{2};
\end{cases}
\quad 
e(\lambda_{*}; s) \simeq 
\begin{cases}
\sigma^{2 - \frac{2\beta}{2r+1}}, &s > r - \frac{\beta + 1}{2},\\
\sigma^{2 - \frac{2\beta}{2s+2+\beta}},  &s < r - \frac{\beta + 1}{2},
\end{cases}
\end{equation}
where $\beta$ is the constant defined in \eqref{eq:beta}. 
\end{theorem}

We introduce a generic small noise analysis scheme to compute the rates. It extends the classical idea of bias-variance trade-off in learning theory \cite{CuckerSmale02,CZ07book} to ill-posed inverse problems. A key component is an integral approximation of the series under Assumption \ref{assump:spectral_decay} to reveal the dominating orders, making it possible to compute the optimal hyper-parameter $\lambda$ through an algebraic equation. 
 The scheme consists of three steps: 
\begin{itemize}[leftmargin=+.58in]  
\item[Step 1:] Reduce the selection of optimal $\lambda$ to solving an algebraic equation;
\item[Step 2:] Estimate the dominating orders of the series for small $\lambda$ by integrals; 
\item[Step 3:] Solve the algebraic equation and compute the optimal rates.     	
\end{itemize}
This scheme was used in \cite{wahba1977practical} to study the rates when the mesh size vanishes. Our innovation is a systematic formulation for the study of small noise limits.

Step 1 follows by separating the variance and bias terms and solving the equation for a critical point, as shown in Lemma \ref{lemma:lambda_eq}.
\begin{lemma}\label{lemma:lambda_eq}
For each $s\geq 0$, the minimizer $\lambda_*:=\argmin{\lambda>0}e(\lambda,s)$ satisfies  
 \begin{equation}\label{eq:opt_lambda_ABs}
\lambda = - \sigma^2  \frac{A'(\lambda;s) }{2B_1(\lambda;s) },  
\end{equation}
where $A'(\lambda,s)$ and $B_1(\lambda,s)$ are series determined by the spectrum and $r$ in Assumption {\rm\ref{assump:spectral_decay}}: 
 \begin{equation*}
\begin{aligned}
-\frac{1}{2}A'(\lambda;s) = \sum_i (\lambda_i^{s + 1} +\lambda)^{-3} \lambda_i^{2s+1},\quad   
B_1(\lambda;s) = \sum_i (\lambda_i^{s + 1} +\lambda)^{-3} \lambda_i^{s+1+2r}\widetilde{p}_i^2. 
\end{aligned} 
\end{equation*}
\end{lemma}
\begin{proof} Rewrite \eqref{eq:mse_s} to separate the variance and bias terms as 
\begin{align}
e(\lambda;s) & = \sigma^2 A(\lambda;s) + \lambda^2 B(\lambda;s),  \text{ where }  \label{eq:error_lambda_s} \\
A(\lambda;s) = \sum_i (\lambda_i^{s + 1} & +\lambda)^{-2} \lambda_i^{2s+1}, 
 \quad  
B(\lambda;s) = \sum_i (\lambda_i^{s + 1} +\lambda)^{-2} \lambda_i^{2r}\widetilde{p}_i^2.  \label{eq:ABs}
\end{align}
Then for each $s$, the minimizer $\lambda_{*}$ must satisfy the equation for a critical point:
\begin{equation*}
0=\frac{d}{d\lambda} e(\lambda;s)  =\sigma^2 A'(\lambda;s) + 2\lambda [ B(\lambda;s) +\frac{\lambda}{2}B'(\lambda;s)]. 
\end{equation*}
Setting $B_1(\lambda;s)= B(\lambda;s) +\frac{\lambda}{2}B'(\lambda;s)$, we obtain Equ.~\eqref{eq:opt_lambda_ABs}. 	The series for $A'$ and $B_1$ follows from \eqref{eq:ABs} and the derivatives of these terms. 
\end{proof}

In Step 2, we estimate the dominating terms of $A'$ and $B_1$ for small $\lambda$ (since it will decay with $\sigma$), which requires estimating the dominating terms in $A, B$ in \eqref{eq:ABs} and $B'$. All these series can be written in the general form of parameterized series:   
\begin{equation*}
 F_s(\lambda; k,\alpha):=   \sum_{i = 1}^\infty (\lambda_i^{s+1} + \lambda)^{-k}\lambda_i^\alpha,
\end{equation*}
where $k \in\{ 2, 3\}$ and $\alpha \in \{(2s+1), (s+1+2r), 2r\}$. For example,  $A(\lambda;s) = F_s(\lambda; 2,2s+1)$ and $B(\lambda;s) = F_s(\lambda; 2,2r)$.  That is,
\begin{align*}
 A(\lambda;s) &= F_s(\lambda; 2,2s+1), \,\, & -\frac{1}{2}A'(\lambda;s) &= F_s(\lambda; 3,2s+1), \\
 B(\lambda;s) & = F_s(\lambda; 2,2r),\,\,   & B'(\lambda;s) &=  F_s(\lambda; 3,2r), \quad   B_1(\lambda,s)  = F_s(\lambda;3,s+1+2r).
 \end{align*}

We provide a general estimation for series $F_s(\lambda; k,\alpha)$ for all $k> 0$, $\alpha\geq 1$ and $s\geq 0$ in Lemma \ref{lem:major_series}, which is of independent interest by itself. The basic idea in its proof is to approximate the series by Riemann sum. In particular, these series estimations provide dominating terms in the series $A, A', B$ and $B_1$ when $\lambda$ is small, as shown in the next lemma. Both lemmas are technical and their proofs can be found in Appendix \ref{sec:append}. 
We will use the big-O notation to denote the approximation error and highlight the dominating terms.  
\begin{lemma}\label{lemma:seiresAB}
Under the Assumption {\rm\ref{assump:spectral_decay}}, as $\lambda \to 0$, we have 
\begin{equation}\label{eq:seriesAB}
\begin{aligned}
A(\lambda;s)    &= C_{A}\lambda^{-\eta_{A}} + O(1),
&\frac{-1}{2}A'(\lambda;s) &= C_{A'}\lambda^{-\eta_A - 1} + O(1), \\ 
B_1(\lambda;s)  &=   \begin{cases}
	    C_{B_1}\lambda^{-\eta_{B}} + O(1), & \\
     	C_{B_1} + O(1), &  
    \end{cases} 
&B(\lambda;s) & = \begin{cases}
	   C_{B}\lambda^{-\eta_{B}} + O(1), &s > r - \frac{\beta + 1}{2}\\
	   C_{B} + O(1),  &s < r - \frac{\beta + 1}{2}    \end{cases}
\end{aligned}
\end{equation}
where $\eta_{A} = \frac{\beta}{s+1}$, $\eta_{B} = \frac{\beta - 2r - 1}{s+1} + 2$. Here $C_{A'}, C_A, C_{B'}, C_{B}>0$ are constants independent of $\lambda$.
\end{lemma}

In Step 3, we prove Theorem \ref{thm:snl_reguEst} by solving the algebraic equations for the dominating terms and computing the rates. 
\begin{proof}[Proof of Theorem \ref{thm:snl_reguEst}]
When $r < s+\frac{\beta + 1}{2}$, by \eqref{eq:opt_lambda_ABs} and the estimates in Lemma \ref{lemma:seiresAB}, 
the minimizer $\lambda_{*}$ satisfies 
\begin{equation*}
   \lambda = 2\sigma^2 \frac{C_{A'}\lambda^{-\eta_A-1} + O(1)}{C_{B_1}\lambda^{-\eta_B} + O(1)},
 \end{equation*}
where $\eta_{A} = \frac{\beta}{s+1}$ and $\eta_{B} = \frac{\beta - 2r - 1}{s+1} + 2$. 
Hence $\eta_A + 2> \eta_A - \eta_B + 2= \frac{ 2r +1}{s+1} >0$ and   
\begin{equation*}
    \lambda^{\eta_A -\eta_B + 2} + O(\lambda^{\eta_A + 2})= 2C_{A'} {C_{B_1}}^{-1}\sigma^2(1 + O(
    \lambda^{\eta_A + 1}))  
\end{equation*}
when $\sigma \to 0$. Solving from the dominating terms, we obtain, with $c = \rbracket{\frac{2C_{A'}}{C_{B_1}}}^{\frac{s+1}{2r+1}}$, 
\begin{equation*}
    \lambda_{*} \simeq c \sigma^{\frac{2s+2}{2r+1}}.  
\end{equation*}
The optimal $\lambda_{*}$, together with \eqref{eq:error_lambda_s} and the estimates in Lemma \ref{lemma:seiresAB}, imply that the minimum of $e(\lambda;s)$ is 
\begin{align*}
     e(\lambda_{*}; s)& = \sigma^2 A(\lambda_{*};s) + \lambda_*^2 B(\lambda_{*s};s) \\
     &  \simeq     \sigma^2\lambda_{*}^{-\eta_A}C_A + \lambda_{*}^{2-\eta_B}C_B \\
 &  =      c^{-\eta_A} \sigma^{2+\frac{2s+2}{2r+1}(-\eta_A)}C_A + c^{2-\eta_B}\sigma^{\frac{2s+2}{2r+1}(2-\eta_B)}C_B. 
\end{align*}
Notice that $  2+\frac{2s+2}{2r+1}(-\eta_A) = \frac{2s+2}{2r+1}(2-\eta_B) = 2 - \frac{\beta}{2r+1}$, hence,
\begin{equation*}
    e(\lambda_{*}; s)  \simeq C \sigma^{2 - \frac{2\beta}{2r+1}},  
\end{equation*}
where $C = c^{-\eta_A}C_A + c^{2-\eta_B}C_B$.

When $r > s+\frac{\beta + 1}{2}$, the optimal hyper-parameter $\lambda_{*}$ satisfies 
 \begin{equation*}
    \lambda = 2\sigma^2 \frac{C_{A'}\lambda^{-1-\eta_A} + O(1)}{C_{B_1} + O(1)}. 
\end{equation*}
As $\sigma \to 0$, notice that $\frac{2}{\eta_A + 2} = \frac{2s+2}{2s+2+\beta}$, we have 
\begin{equation*}
    \lambda_{*} \simeq c \sigma^{\frac{2s+2}{2s+2+\beta}} 
\end{equation*}
with $c = \rbracket{\frac{2C_{A'}}{C_{B_1}}}^{\frac{s+1}{2s+2+\beta}}$. Hence, 
\begin{align*}
      e(\lambda_{*}; s) =   c^{-\eta_A} \sigma^{2+\frac{2s+2}{2s+2+\beta}(-\eta_A)}C_A + c^{2-\eta_B}\sigma^{\frac{4s+4}{2s+2+\beta}}C_B. 
\end{align*}
Again, noticing that $  2+\frac{2s+2}{2s+2+\beta}(-\eta_A) = \frac{4s+4}{2s+2+\beta} = 2 - \frac{\beta}{2s+2+\beta}$, we obtain
\[
e(\lambda_{*}; s)   = C \sigma^{2 - \frac{\beta}{2s+2+\beta}},
\]
where $C = c^{-\eta_A}C_A + c^{2-\eta_B}C_B$. 
\end{proof}
\begin{remark}\label{rmk:rate-threshold}
 The convergence rate at the threshold $s=r-\frac{\beta+1}{2}$ is not covered by Theorem \ref{thm:snl_reguEst}, because it requires a solution of a non-algebraic equation involving logarithmic terms.  
\end{remark}



\section{Numerical Examples} \label{sec:num}

\subsection{Oracle rates under exponential spectral decay} \label{sec:oracle-rate}
We examine first the optimal convergence rate when the hyper-parameter is selected to minimize the $L^2_\rho$ error, as in Theorem \ref{thm:snl_reguEst}. Since the true function is used in the selection of the optimal hyper-parameter, we call this rate the oracle rate.

The setting is as follows. We take the spectrum to be $f(x) = e^{-1.5 x}$ 
evaluated at $i\in \{1,\ldots,200\}$, and $r \in \{0.7, 1.2, 1.7\}$. For each $s$, we select the optimal $\lambda_*$ to be the minimizer of $e(\lambda;s)$ in \eqref{eq:mse_s} in the range $[10^{-25}, 10^{2}]$ with true $\phi$ known, and then compute $e(\lambda_*;s)$. We compute the rate of $\lambda_*$ and $e(\lambda_*; s)$ by linear regression from their values for a sequence of $\sigma \in [10^{-7}, 10^{-1}]$. 

 Fig.~\ref{fig:thm_apx} demonstrates the dependence of optimal convergence rate on $s$ and $r$. In the left figure, the theoretical rates (denoted by ``Theoretical'') in Theorem \ref{thm:snl_reguEst} are close to the numerically approximated rates (``Approximated''). They differ the most near the thresholds $r - \frac{\beta + 1}{2}$, which is not covered by the theorem (see Remark \ref{rmk:rate-threshold}). 
Both show that over-smoothing leads to optimal convergence rates. However, as the right figure shows, the optimal hyperparameter decays at a rate increasing with $s$ linearly, making it difficult to select in practice, particularly when $s$ is relatively large. The example in the next section will show that over-smoothing leads to large oscillatory hyper-parameters in practice.

\subsection{Rates in practice under polynomial spectral decay.}
 We further investigate the effect of over-smoothing and under-smoothing in a practical application setting. In practice, the ground truth $\phi_*$ is unknown, and one has to use empirical methods to select the optimal hyper-parameter $\lambda$. In our tests, the L-curve method in \cite{hansen_LcurveIts_a} performs similarly to the generalized cross-validation (GCV) method in \cite{golub1979generalized}, and we present only the results of the L-curve method.

 We consider the Fredholm integral equation \eqref{eq:FIE} with kernel 
\begin{equation}\label{eq:K}
 K_1(x, y) =
    \begin{cases}
        -\frac{\sin 2(1-y)\sin 2x}{2\sin 2},\ 0 \leq x \leq y \leq 1 ,   \\
        -\frac{\sin 2(1-x)\sin 2y}{2\sin 2},\ 0 \leq y \leq x \leq 1 .   \\
    \end{cases}
\end{equation}
Note that $K_1$ is the Green's function of the equation 
\begin{equation*}
    u(x)'' + 4u(x)  = f(x), \ x\in (0, 1), \ u(0)  = u(1) = 0.
  \end{equation*}

We take $\rho$ to be the Lebesgue measure on $[0, 1]$ and set $\spaceY = L^2([0, 1]) = L^2_\rho$. 
 As introduced in \eqref{least_sq_org}, the loss functional is given by
\begin{equation*}
    \mathcal{E}(\phi) = \innerp{L_K^*L_K\phi, \phi}_{L^2_\rho} - 2\innerp{L_K^* y, \phi}_{L^2_\rho} + \norm{y}_{L^2_\rho}^2 = \innerp{\LGbar\phi, \phi}_{L^2_\rho} - 2\innerp{\phi^\by, \phi}_{L^2_\rho} + C_\by, 
\end{equation*}
where the integral kernel is 
 \begin{equation*}
    \Gbar(x, z) = \int_{0}^1 K(x, w)K(w, z) dw.  
\end{equation*}

Direct computation shows that $\LGbar$ has eigenvalues $2(4 - n^2\pi^2)^{-2}$ with eigenfunctions $\phi_n(x) = \sin(n\pi x)$. That is, its spectrum has polynomial decay with $\theta=4$, so $\beta=\frac{5}{4}$. 
We evaluate the kernel in $[0, 1]$ with $500$ even-spaced mesh points and take the true function $\phi = \sum_{n = 1}^N v_n\lambda_n^r\phi_n$ with $r = 1.5$ and $\{v_n\}$  uniformly sampled in $[-1.05, -0.95]\cup[0.95, 1.05]$. We estimate $\phi$ from noisy observations with $\sigma \in[10^{-3}, 10^{-0.5}]$ and $H_G^s$-regularization with $s \in \{0, 1, 2\}$. We test two approaches estimating $\lambda_*$:  using the L-curve method in  \cite{hansen_LcurveIts_a} (denoted by ``L-curve $\lambda_*$") and using the true function $\phi$ (denoted by ``oracle $\lambda_{*}$"). 

\begin{figure}
    \centering
\includegraphics[width=1\textwidth]{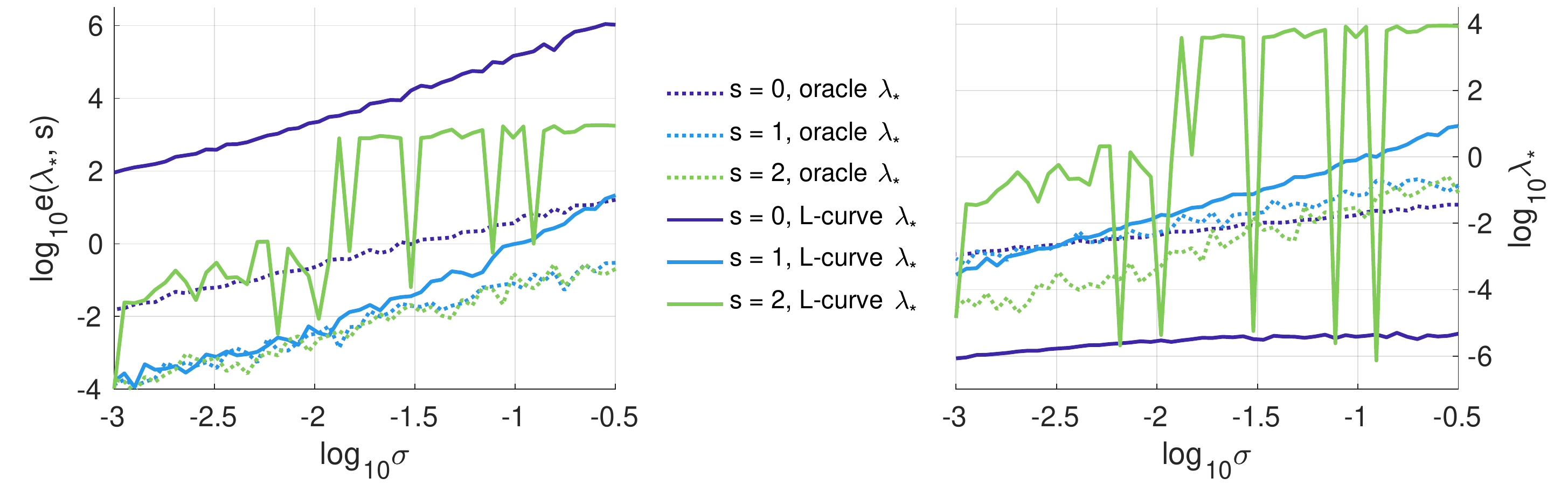}
    \caption{over-smoothing ($s=2$) makes it difficult to select the optimal $\lambda_*$ by the L-curve method (right), leading to a relatively large error (left). 
However, when properly regularized with $s=1$, the L-curve $\lambda_*$'s are close to the oracle ones and lead to estimators that are significantly more accurate than those with $s=0$ or $s=2$.   
 }
     \label{fig:FIE} 
\end{figure}

Fig.~\ref{fig:FIE} shows the errors (Left) and the estimated hyper-parameters (Right) as $\sigma$ decays. As $s$ increases, the errors with L-curved estimated $\lambda_*$ (solid lines) first drop then increase, whereas the errors with oracle $\lambda_*$ decay steadily.   In particular, for the over-smoothed case with $s=2$, the optimal $\lambda_*$ selected by the L-curve method is unstable  (Right), leading to large oscillating errors. Meanwhile, for the under-smoothed case with $s=0$, the L-curve selects hyper-parameters that are much smaller than the optimal ones, leading to under-regularized estimators.  
 
However, for the properly-smoothed regularization with $s=1$, the L-curve $\lambda_*$'s are close to the oracle ones and lead to estimators that are significantly more accurate than those with $s=0$ or $s=2$. Note that the threshold is $r - \frac{\beta +1}{2} = 0.375$. Hence $s = 1$ is big enough to have the optimal error rate.

In short, this test suggests that as the noise level decays, an over-smoothing regularizer tends to lead to oscillatory hyper-parameters, and an under-smoothing regularizer tends to lead to flat hyper-parameters, both produce inaccurate estimators. In contrast, a properly smooth regularizer leads to near-optimal hyper-parameters and accurate estimators that decay with the noise level.

\section{Limitations}  
The small noise analysis does not tackle the practical selection of hyper-parameters. Its goal is to understand the fundamental role of regularization norms and to shed insights. Thus, it uses the oracle optimal hyper-parameters so as to focus on comparing the norms. It remains open to investigate the joint selection of regularization norms and hyper-parameters without knowing the true solution in practice. 
 

Also, the small noise analysis does not apply to non-Hilbert norms, particularly the $L^1$-norm; it is potentially applicable to other Hilbert norms, such as the $H_0^1$-norm and those non-diagonalizable in the orthonormal basis of the normal operator, which we leave in future work.

Additionally, it is yet to extend this framework to analyze the convergence of the regularized estimator when the data size increases (i.e., either the sample size increases or the data mesh increases).

Lastly, this study focuses on linear inverse problems. Extending it to nonlinear problems requires varying second-order Fr'echet derivatives of the loss function, which is beyond the current scope.
 
\section{Conclusion}\label{sec:conlusion} 
We have established a small noise analysis framework to understand the fundamental role of regularization norms, setting a step forward for the anatomy of regularization.  
 For ill-posed linear inverse problems, we have introduced a class of adaptive fractional Sobolev space norms for regularization. Our analysis shows that over-smoothing via fractional spaces always leads to an optimal convergence rate, but the resulting fast-decaying optimal hyper-parameter becomes difficult to select in practice.  

\appendix

\section{Technical proofs}\label{sec:append}
\begin{proof}[Proof of Lemma \ref{lemma:LSE}]  We first show that the estimator in \eqref{eq:LSE} is the unique minimizer of the loss function $\calE$ in $H$, i.e., it is the unique zero of the Fr\'echet derivative of $\calE$ in $H$. By \eqref{eq:lossFn} and \eqref{eq:LGbar}--\eqref{eq:LSE}, we can write the loss functional as 
\begin{equation}\label{eq:loss_L2}
\calE(\phi) =  \innerp{\LGbar \phi,\phi}_{L^2_\rho} +  \innerp{\phi^\by,\phi}_{L^2_\rho} + C_\by 
\end{equation}
with $C_\by =\frac{1}{n}\sum_{k=1}^n\|y_k\|_\spaceY^2$. 
Then, the Fr\'echet derivative $\nabla \calE(\phi)$ in $L^2_\rho$ is 
\beq \label{eq:Frechet}
\innerp{\nabla \calE(\phi), \psi}_{L^2_\rho} = \lim_{h\to 0} \frac{\calE(\phi+h\psi) - \calE(\phi)}{h} =  \innerp{2 ( \LGbar \phi -\phi^\by), \psi}_{L^2_\rho}, \, \forall \psi \in L^2_\rho, 
\eeq
that is, $\nabla \calE(\phi) = 2 ( \LGbar \phi -\phi^\by)$. Hence, by the definition of $H$, the estimator in \eqref{eq:LSE} is the unique zero of $\nabla \calE(\phi)$ in $H$. Additionally, note that any LSE can be written as $\phi_{LSE} = \widehat{\phi}+ \phi_{null}\in H\oplus \mathrm{Null}(\LGbar)$. Thus, the estimator $\widehat{\phi}$ is the mini-norm LSE. 

Next, we prove the estimator's properties.  \\
\textbf{Part (a):} By the definition of $\phi^\by$, we have   
\begin{align*}
\phi^\by =\frac{1}{n}\sum_{k=1}^n L_k^* (L_k\phi_*+\sigma \dot W_k) = \LGbar \phi_* + \phi^\sigma, \quad \phi^\sigma= \sigma \frac{1}{n}\sum_{k=1}^n L_k^* \dot W_k. 
\end{align*}
The distribution of $\phi^\sigma$ is $\mathcal{N}(0,\sigma^2\LGbar)$ because for each $\psi \in L^2_\rho$, the random variable $ \innerp{\psi,\phi^\sigma}_{L^2_\rho} =  \frac{1}{n}\sum_{k=1}^n  \innerp{L\psi,\sigma \dot W_k}_\spaceY$ is Gaussian with mean zero and variance $\sigma^2\innerp{\psi,\LGbar\psi}_{L^2_\rho}$ . Therefore, we can write $\phi^\sigma= \sum_i \sigma \xi_i \lambda_i^{1/2}\psi_i $ with $\{\xi_i\}$ being i.i.d.~standard Gaussian, and $\Ebracket{\|\phi^\sigma\|_{L^2_\rho}^2}= \sigma^2\sum_i\lambda_i$, where the sum is finite by Assumption \ref{assumption:LG}.  \\ 
\textbf{Part (b):}  note that  $\LGbar^{-1}\phi^\sigma = \sum_{i:\lambda_i>0} \lambda_i^{-1/2} \sigma \xi_i \psi_i $. 
Then,  when $\sum_{i} \lambda_i^{-1}<\infty$ (which happens only when there are finitely many non-zero eigenvalues), we have $ \LGbar^{-1}\phi^\sigma \in L^2_\rho$ because $\Ebracket{ \| \LGbar^{-1}\phi^\sigma\|_{L^2_\rho}^2} =\sigma^2  \sum_{i} \lambda_i^{-1}<\infty$. Then,  $\LGbar^{-1}\phi^\sigma\sim \calN(0,\sigma^2\LGbar^{-1})$ and the distribution of the estimator follows. 
But when $\sum_{i} \lambda_i^{-1}=\infty$, we have $\Ebracket{\| \LGbar^{-1}\phi^\sigma \|_{L^2_\rho}^2 }
 = \sum_{i} \lambda_i^{-1} \sigma^2 
=\infty$, and hence $\LGbar^{-1}\phi^\by$ is ill-defined.\\
\textbf{Part (c):} When the data is noiseless,  we have $\phi^\by= \LGbar\phi_* $. Hence $\widehat{\phi} = \LGbar^{-1}\phi^\by = P_H \phi_*$. 
\end{proof}

\begin{lemma}[Series estimation]\label{lem:major_series}
Assume that the sequence $\lambda_i$ satisfies \eqref{eq:decay_f}, where $f(x) = e^{-\theta (x-1)}$ or $f(x) = x^{-\theta}$ with $\theta > 1$. Then, for all $k > 0$, $\alpha \geq 1$ and  $s \geq 0$, as $\lambda \to 0$,  we have 
\begin{equation} \label{eq:Fsk-alapha}
    F_s(\lambda; k,\alpha)= \sum_{i = 1}^\infty (\lambda_i^{s+1} + \lambda)^{-k}\lambda_i^\alpha = CJ(\lambda) +O(1),
\end{equation}
where  $C = \frac{C_1^{(s+1)k}C_2^{-\alpha}}{\theta}$ with constants $C_1, C_2 \in [b^{-1}, a^{-1}]$ and 
\begin{equation}\label{eq:Jlambda}
        J(\lambda) \simeq
        \begin{cases}
            \frac{c^{\gamma - k}}{s+1}\frac{\Gamma(\gamma)\Gamma(k - \gamma)}{\Gamma(k)} \lambda^{\gamma - k}, &0 < \gamma < k;\\
            \frac{1}{s+1} \ln(1/\lambda), &\gamma = k;\\
            \frac{1}{s+1}\frac{1}{\gamma - k}, &\gamma > k,
        \end{cases}
\end{equation}
where $\gamma = \frac{\alpha - \beta +1}{s+1}$, $c = C_1^{s+1}$, and $\beta$ is defined in \eqref{eq:beta}. 
Here "$\simeq$" indicates an approximate equation up to a difference of $O(\lambda)$ as $\lambda \to 0$.  
\end{lemma}

To prove Lemma \ref{lem:major_series}, we introduce first an intermediate value theorem for series. It shows that the bounded multiplicative perturbations of the eigenvalues $\{\lambda_i\}$ have negligible effects on the series involving the sequence.    
\begin{lemma}[Intermediate value theorem with series]\label{lemma:seiresMVT}
Let $0<c_1 < c_2<\infty$, $\lambda\geq 0$, and $k\geq 1$. 
Let $\{f_i,g_i\}_{i=1}^n$ be two non-negative sequences such that $\sum_{i=1}^n (c_1 g_i +\lambda )^{-k} c_2 f_i < \infty$ where $n\leq \infty$. Then, for any sequence $\{b_i\}$ such that $c_1\leq b_i\leq c_2$ for all $i$, there exist constants $C_1,C_2 \in [c_1,c_2]$ such that  
\[
\sum_{i=1}^n (b_i g_i + \lambda)^{-k} b_i f_i =  \sum_{i=1}^n (C_1 g_i + \lambda)^{-k} C_2 f_i. 
\]     	
\end{lemma}
\begin{proof}
Without loss of generality, we consider only the case when $n=\infty$, since the proof below applies to the case when $n<\infty$, and the constants $C_1,C_2$ are always in the interval $[c_1,c_2]$ regardless of $n$.  

Let $h(x) = \sum_{i=1}^\infty (x g_i +\lambda )^{-k} b_i f_i$, where the series converges for each $x \in [c_1,c_2]$ since $(x g_i +\lambda )^{-k} b_i f_i\leq (c_1 g_i +\lambda )^{-k} c_2 f_i$ for each $i$ and  $\sum_{i=1}^\infty (c_1 g_i +\lambda )^{-k} c_2 f_i < \infty$. Note that $h$ is continuous. Since $h(c_2)\leq \sum_{i=1}^\infty (b_i g_i + \lambda)^{-k} b_i f_i \leq h(c_1)$, we obtain by the intermediate value theorem that there exists $C_1\in [c_1,c_2]$ such that 
\[ \sum_{i=1}^\infty (b_i g_i + \lambda)^{-k} b_i f_i = h(C_1)= \sum_{i=1}^\infty (C_1 g_i + \lambda)^{-k} b_i f_i.
\] 

Similarly, let $h_{C_1}(x) =\sum_{i=1}^\infty (C_1 g_i + \lambda)^{-k} x f_i $, where the series converges for any $x\in [c_1,c_2]$. Note that $h_{C_1}$ is continuous with $h_{C_1}(c_1)\leq \sum_{i=1}^\infty (C_1 g_i + \lambda)^{-k} b_i f_i \leq h_{C_1}(c_2)$. Applying the intermediate value theorem again, we see that there exists $C_2\in [c_1,c_2]$ such that 
\[ \sum_{i=1}^\infty (C_1 g_i + \lambda)^{-k} b_i f_i = h_{C_1}(C_2)= \sum_{i=1}^\infty (C_1 g_i + \lambda)^{-k} C_2 f_i.
\]

Combining the above two equations, we conclude the proof. 
\end{proof}

\begin{proof}[Proof of Lemma \ref{lem:major_series}]
The proof is based on approximating the series by the corresponding Riemann integral. The approximation error is of order $O(1)$, which is negligible when compared to a negative power of small $\lambda$.

    Firstly, by \eqref{eq:decay_f} and the mean-value theorem in Lemma \ref{lemma:seiresMVT}, there exists $C_1, C_2 \in [b^{-1}, a^{-1}]$ such that 
    \begin{align*}
        F_s(\lambda; k,\alpha) 
        &= \sum_{i = 1}^\infty (p_i^{-(s+1)}f(i)^{s+1} + \lambda)^{-k}(p_i^{-1}f(i))^\alpha \\
       & = \sum_{i = 1}^\infty (C_1^{-(s+1)}f(i)^{s+1} + \lambda)^{-k}(C_2^{-1}f(i))^\alpha.
    \end{align*}
 Then, approximating the series by integral, and using the notation $c=C_1^{s+1}$, we have 
    \begin{equation*}
        \sum_{i = 1}^\infty (c^{-1}f(i)^{s+1} + \lambda)^{-k}(C_2^{-1}f(i))^\alpha = c^{k}C_2^{-\alpha} \int_1^\infty (f(x)^{s+1} + c\lambda)^{-k} f(x)^\alpha dx + O(1). 
    \end{equation*}
    We take $g = f^{-1}$ so that $y = f(x)$ and $g(y) = x$. Then, we have $dx = g'(y) dy$ and 
    \begin{align*}
        \int_1^\infty (f(x)^{s+1} + c\lambda)^{-k} f(x)^\alpha dx 
        = 
        \int_1^0 (y^{s+1} + c\lambda)^{-k} y^{\alpha} g'(y) dy, 
    \end{align*}
    where we used the fact that $f(1)=1$ and $\lim_{x\to\infty} f(x) =0$ when $f$ has either exponential or polynomial decay. 
    
    Before continuing the computation, we introduce $\beta$ in \eqref{eq:beta} to unify the notation for both cases of $f$. 
     If $f(x) = e^{-\theta(x-1)}$, then $g(y) = -\frac{1}{\theta}\ln(y)+1$, so that $g'(y) = -\frac{1}{\theta}y^{-1}$. If $f(x) = x^{-\theta}$, then $g(y) = y^{-1/\theta}$, so that $g'(y) = -\frac{1}{\theta}y^{-\frac{1}{\theta}-1}$. In either case, we can assume $g'(y) = -\frac{1}{\theta} y^{-\beta}$, where $\beta = 1$ or $\beta = \frac{1}{\theta} + 1$. Then we have
    \begin{align*}
        \int_1^0 (y^{s+1} +c\lambda)^{-k} y^{\alpha} g'(y) dy
        = \frac{1}{\theta}\int_0^1 (y^{s+1} +c\lambda)^{-k} y^{\alpha - \beta} dy. 
    \end{align*}
    
  Next, we estimate the above integral, denoted by $J(\lambda)$: 
\[     J(\lambda) = \int_0^1 (y^{s+1} + c\lambda)^{-k} y^{\alpha - \beta} dy
\] 
in three cases: $\gamma>k$, $0<\gamma<k$ and $\gamma=k$ (notice that $\gamma = \frac{\alpha - \beta +1}{s+1}$ is positive from the range of $\alpha$ and $\beta$).

 When $\gamma > k$, we have 
    \begin{equation*}
        \lim_{\lambda \to 0} J(\lambda) = \int_{0}^1 y^{-k(s+1) + \alpha - \beta} dy = \frac{1}{(s+1)(\gamma - k)}, 
    \end{equation*}
    which proves the last row in \eqref{eq:Jlambda}. 
 To tackle the cases $\gamma\leq k$, we make another change of variable $z = \frac{y^{s+1}}{c\lambda}$, so that $y = (c\lambda z)^{\frac{1}{s+1}}$, $dy = \frac{1}{s+1}(c\lambda)^{\frac{1}{s+1}}z^{\frac{1}{s+1} - 1} dz$, and then
    \begin{align*}
        J(\lambda) &= \frac{1}{s+1}(c\lambda)^{\frac{\alpha - \beta + 1}{s+1} - k}\int_0^{1/(c\lambda)} (z + 1)^{-k}z^{\frac{\alpha - \beta +1}{s+1} - 1}dz  \\
       & = \frac{1}{s+1}(c\lambda)^{\gamma - k} I(\lambda), \quad I(\lambda):=\int_0^{1/(c\lambda)} (z + 1)^{-k}z^{\gamma - 1}dz.  
    \end{align*}

    When $0 < \gamma < k$, we have 
    \begin{equation*}
        \lim_{\lambda \to 0} I(\lambda) = \int_0^\infty (z+1)^{-k}z^{\gamma - 1}dz = \frac{\Gamma(\gamma)\Gamma(k - \gamma)}{\Gamma(k)}.
    \end{equation*}
    When $\gamma = k$, we notice that $I(\lambda) \to \infty$ as $\lambda \to 0$. By L'Hospital rule, 
    \begin{equation}
        \lim_{\lambda \to 0}\frac{I(\lambda)}{\ln(1/\lambda)} = \lim_{\gamma \to 0}\frac{I'(\lambda)}{-(1/\lambda)}
        = 
        \lim_{\gamma \to 0}\frac{(c\lambda + 1)^{-k} \lambda^{-1}}{\lambda^{-1}} = 1.
    \end{equation}
    
    Combining the above equations, we obtain \eqref{eq:Jlambda} and \eqref{eq:Fsk-alapha}
\end{proof}

\begin{proof}[Proof of Lemma \ref{lemma:seiresAB}] The estimations follow by applying Lemma \ref{lem:major_series}. 

    The estimation for $A(\lambda;s) = F_s(\lambda; 2,2s+1)$ follows from Lemma \ref{lem:major_series} with $k=2,\alpha=2s+1$: since $\gamma =\frac{\alpha - \beta +1}{s+1}= 2 -\frac{\beta}{s+1} < k$, we use the first case of $J(\lambda)$ in \eqref{eq:Jlambda} to obtain $A(\lambda;s) = C_{A}\lambda^{-\eta_{A}} + O(1)$ with $\eta_A = k - \gamma = \frac{\beta}{s+1}$.

    For $\frac{-1}{2}A'(\lambda;s)$, we have $\alpha = 2s+1, k = 3$, hence $\gamma = \frac{\alpha - \beta + 1}{s+1} = 2 - \frac{\beta}{s+1} < 3$. Then
    \begin{equation*}
       \frac{-1}{2}A'(\lambda;s) = C_{A'}\lambda^{-\eta_{A} - 1} + O(1). 
    \end{equation*}
    
    To estimate $B_1(\lambda;s)$ and $B(\lambda;s)$, we note that the perturbation terms $\{\widetilde{p}_i\}$ in $B_1(\lambda;s)$ and $B(\lambda;s)$ are bounded above and below by Assumption \ref{assump:spectral_decay}. Thus, these perturbation terms do not affect the series estimation, which follows from the same arguments the proof of Lemma \ref{lem:major_series} and Lemma \ref{lemma:seiresMVT}. 
    
    Thus,  
    For $B_1(\lambda;s)$, we have $\alpha = s+1+2r$, $k = 3$ and $\gamma = \frac{1 + 2r - \beta}{s+1}+ 1$. For $B(\lambda;s)$, we have $\alpha = 2r$, $k = 2$ and $\gamma = \frac{1 + 2r - \beta}{s+1}$. Hence when $\frac{1 + 2r - \beta}{s+1} < 2$, namely $r < s+\frac{\beta + 1}{2}$, we have $\gamma<k$, so  
    \begin{equation*}
    \begin{aligned}
    B_1(\lambda;s)  = C_{B_1}\lambda^{-\eta_{B}} + O(1), \ 
    B(\lambda;s)    = C_{B}\lambda^{-\eta_{B}} + O(1),
    \end{aligned}
    \end{equation*}
    where $\eta_{B} =2-\gamma = \frac{\beta - 2r - 1}{s+1} + 2$. When $\frac{1 + 2r - \beta}{s+1} > 2$, namely $r > s+\frac{\beta + 1}{2}$, we have 
    \begin{equation*}
    \begin{aligned}
    B_1(\lambda;s)  = C_{B_1} + O(1), \ 
    B(\lambda;s)    = C_{B} + O(1).
    \end{aligned}
    \end{equation*}
    Note that the constants $C_B$, $C_{B_1}$ might be different in the two cases. 
\end{proof}

\paragraph{Acknowledgments}{The authors would like to thank Dr.~Tao Zhou for his helpful comments. FL is partially supported by the Johns Hopkins University Catalyst Award, the NSF DMS- 2238486, and AFOSR FA9550-21-1-0317. }

\bibliographystyle{plain}

\begin{thebibliography}{10}

\bibitem{balasubramanian2022unified}
Krishnakumar Balasubramanian, Hans-Georg M{\"u}ller, and Bharath~K
  Sriperumbudur.
\newblock Unified {RKHS} methodology and analysis for functional linear and
  single-index models.
\newblock {\em arXiv preprint arXiv:2206.03975}, 2022.

\bibitem{bauer2007regularization}
Frank Bauer, Sergei Pereverzev, and Lorenzo Rosasco.
\newblock On regularization algorithms in learning theory.
\newblock {\em Journal of complexity}, 23(1):52--72, 2007.

\bibitem{belge2002efficient}
Murat Belge, Misha~E Kilmer, and Eric~L Miller.
\newblock Efficient determination of multiple regularization parameters in a
  generalized l-curve framework.
\newblock {\em Inverse problems}, 18(4):1161, 2002.

\bibitem{buckwar2010stochastic}
Evelyn Buckwar, Andreas R{\"o}{\ss}ler, and Renate Winkler.
\newblock {Stochastic Runge--Kutta methods for It{\^o} SODEs with small noise}.
\newblock {\em SIAM Journal on Scientific Computing}, 32(4):1789--1808, 2010.

\bibitem{chada2022data}
Neil~K Chada, Quanjun Lang, Fei Lu, and Xiong Wang.
\newblock A data-adaptive prior for {Bayesian} learning of kernels in
  operators.
\newblock {\em arXiv preprint arXiv:2212.14163}, 2022.

\bibitem{chen2022stochastic}
Zhiming Chen, Wenlong Zhang, and Jun Zou.
\newblock Stochastic convergence of regularized solutions and their finite
  element approximations to inverse source problems.
\newblock {\em SIAM Journal on Numerical Analysis}, 60(2):751--780, 2022.

\bibitem{CuckerSmale02}
Felipe Cucker and Steve Smale.
\newblock On the mathematical foundations of learning.
\newblock {\em Bulletin of the American mathematical society}, 39(1):1--49,
  2002.

\bibitem{CZ07book}
Felipe Cucker and Ding~Xuan Zhou.
\newblock {\em Learning theory: an approximation theory viewpoint}, volume~24.
\newblock Cambridge University Press, {Cambridge}, 2007.

\bibitem{da2014stochastic}
Giuseppe Da~Prato and Jerzy Zabczyk.
\newblock {\em Stochastic equations in infinite dimensions}.
\newblock Cambridge university press, 2014.

\bibitem{dashti2013map}
Masoumeh Dashti, Kody~JH Law, Andrew~M Stuart, and Jochen Voss.
\newblock Map estimators and their consistency in bayesian nonparametric
  inverse problems.
\newblock {\em Inverse Problems}, 29(9):095017, 2013.

\bibitem{engl1996regularization}
Heinz~Werner Engl, Martin Hanke, and Andreas Neubauer.
\newblock {\em Regularization of inverse problems}, volume 375.
\newblock Springer Science \& Business Media, 1996.

\bibitem{gazzola2019ir}
Silvia Gazzola, Per~Christian Hansen, and James~G Nagy.
\newblock Ir tools: a matlab package of iterative regularization methods and
  large-scale test problems.
\newblock {\em Numerical Algorithms}, 81(3):773--811, 2019.

\bibitem{golub1979generalized}
Gene~H Golub, Michael Heath, and Grace Wahba.
\newblock Generalized cross-validation as a method for choosing a good ridge
  parameter.
\newblock {\em Technometrics}, 21(2):215--223, 1979.

\bibitem{goodman2016small}
Jonathan Goodman, Kevin~K Lin, and Matthias Morzfeld.
\newblock Small noise analysis and symmetrization of implicit {Monte Carlo}
  samplers.
\newblock {\em Communications on Pure and Applied Mathematics},
  69(10):1924--1951, 2016.

\bibitem{Gray06}
Robert~M Gray.
\newblock {\em Toeplitz and circulant matrices: {A} review}.
\newblock now publishers inc, 2006.

\bibitem{hadamard1923lectures}
Jacques~Salomon Hadamard.
\newblock {\em Lectures on Cauchy's problem in linear partial differential
  equations}, volume~18.
\newblock Yale University Press, 1923.

\bibitem{hall2007methodology}
Peter Hall and Joel~L Horowitz.
\newblock Methodology and convergence rates for functional linear regression.
\newblock {\em Annals of Statistics}, 35(1):70--91, 2007.

\bibitem{hansen1998rank}
Per~Christian Hansen.
\newblock {\em Rank-deficient and discrete ill-posed problems: numerical
  aspects of linear inversion}.
\newblock SIAM, 1998.

\bibitem{hansen_LcurveIts_a}
Per~Christian Hansen.
\newblock The {L}-curve and its use in the numerical treatment of inverse
  problems.
\newblock In {\em in Computational Inverse Problems in Electrocardiology, ed.
  P. Johnston, Advances in Computational Bioengineering}, pages 119--142. WIT
  Press, 2000.

\bibitem{LangLu22}
Quanjun Lang and Fei Lu.
\newblock Learning interaction kernels in mean-field equations of first-order
  systems of interacting particles.
\newblock {\em SIAM Journal on Scientific Computing}, 44(1):A260--A285, 2022.

\bibitem{Li2005modified}
Gongsheng Li and Zuhair Nashed.
\newblock A modified {Tikhonov} regularization for linear operator equations.
\newblock {\em Numerical functional analysis and optimization},
  26(4-5):543--563, 2005.

\bibitem{LAY22}
Fei Lu, Qingci An, and Yue Yu.
\newblock Nonparametric learning of kernels in nonlocal operators.
\newblock {\em Journal of Peridynamics and Nonlocal Modeling}, pages 1--24,
  2023.

\bibitem{LLA22}
Fei Lu, Quanjun Lang, and Qingci An.
\newblock {Data adaptive RKHS Tikhonov regularization for learning kernels in
  operators}.
\newblock {\em Proceedings of Mathematical and Scientific Machine Learning,
  PMLR 190:158-172}, 2022.

\bibitem{LuOu23}
Fei Lu and Miao-Jung~Yvonne Ou.
\newblock An adaptive {RKHS regularization for Fredholm} integral equations.
\newblock {\em arXiv preprint arXiv:2303.13737}, 2023.

\bibitem{nashed1974generalized}
M~Zuhair Nashed and Grace Wahba.
\newblock Generalized inverses in reproducing kernel spaces: an approach to
  regularization of linear operator equations.
\newblock {\em SIAM Journal on Mathematical Analysis}, 5(6):974--987, 1974.

\bibitem{rudin1992nonlinear}
Leonid~I Rudin, Stanley Osher, and Emad Fatemi.
\newblock Nonlinear total variation based noise removal algorithms.
\newblock {\em Physica D: nonlinear phenomena}, 60(1-4):259--268, 1992.

\bibitem{scampicchio2023error}
Anna Scampicchio, Elena Arcari, and Melanie~N Zeilinger.
\newblock Error analysis of regularized trigonometric linear regression with
  unbounded sampling: a statistical learning viewpoint.
\newblock {\em arXiv preprint arXiv:2303.09206}, 2023.

\bibitem{smale2007learning}
Steve Smale and Ding-Xuan Zhou.
\newblock Learning theory estimates via integral operators and their
  approximations.
\newblock {\em Constructive approximation}, 26(2):153--172, 2007.

\bibitem{tihonov1963solution}
Andrei~Nikolajevits Tihonov.
\newblock Solution of incorrectly formulated problems and the regularization
  method.
\newblock {\em Soviet Math.}, 4:1035--1038, 1963.

\bibitem{tikhonov1995numerical}
Andrei~Nikolaevich Tikhonov, AV~Goncharsky, Vyacheslav~Vasil'evich Stepanov,
  and Anatoly~G Yagola.
\newblock {\em Numerical methods for the solution of ill-posed problems},
  volume 328.
\newblock Springer Science \& Business Media, 1995.

\bibitem{wahba1973convergence}
Grace Wahba.
\newblock Convergence rates of certain approximate solutions to fredholm
  integral equations of the first kind.
\newblock {\em Journal of Approximation Theory}, 7(2):167--185, 1973.

\bibitem{wahba1977practical}
Grace Wahba.
\newblock Practical approximate solutions to linear operator equations when the
  data are noisy.
\newblock {\em SIAM journal on numerical analysis}, 14(4):651--667, 1977.

\bibitem{wen2011regularized}
Jin Wen and Ting Wei.
\newblock {Regularized solution to the {Fredholm} integral equation of the
  first kind with noisy data}.
\newblock {\em Journal of applied mathematics \& informatics}, 29(1\_2):23--37,
  2011.

\bibitem{youYu2022_Datadriven}
Huaiqian You, Yue Yu, Stewart Silling, and Marta D’Elia.
\newblock A data-driven peridynamic continuum model for upscaling molecular
  dynamics.
\newblock {\em Computer Methods in Applied Mechanics and Engineering},
  389:114400, 2022.

\bibitem{yuan_cai2010}
Ming Yuan and T~Tony Cai.
\newblock {A reproducing kernel Hilbert space approach to functional linear
  regression}.
\newblock {\em The Annals of Statistics}, 38(6):3412--3444, 2010.

\bibitem{zhang2022stochastic}
Ye~Zhang and Chuchu Chen.
\newblock Stochastic asymptotical regularization for linear inverse problems.
\newblock {\em Inverse Problems}, 39(1):015007, 2022.

\end{thebibliography}

\end{document}